\documentclass{article}
\usepackage{ijcai19}

\usepackage{times}
\usepackage{soul}
\usepackage{url}
\usepackage[hidelinks]{hyperref}
\usepackage[utf8]{inputenc}
\usepackage[small]{caption}
\usepackage{graphicx}
\usepackage{amsmath}
\usepackage{booktabs}
\usepackage{algorithm}
\usepackage[usenames]{color} 
\usepackage{amsthm}
\usepackage{tikz}
\usepackage{dsfont}
\usepackage{units}
\newcommand{\citet}[1]
  {\citeauthor{#1}?\shortcite{#1}}
  \newcommand{\citep}{\cite}
  
\usepackage{listings}
\usepackage[noend]{algpseudocode}
\usepackage{chessboard}
\MakeRobust{\Call}
\usepackage{graphicx}
\graphicspath{ {images/} }

\makeatletter
\def\BState{\State\hskip-\ALG@thistlm}
\makeatother

\usepackage{array}
\usepackage{arydshln}
\newtheorem{example}{Example}

\title{Similarity Measures based on Local Game Trees}

\author{
Sabrina Evans$^1$
\and
Paolo Turrini$^2$\footnote{Contact Author} \\
\affiliations
$^1$ Department of Mathematics, Yale University\\
$^2$ Department of Computer Science, University of Warwick\\
\emails
sabrina.evans@yale.edu, 
p.turrini@warwick.ac.uk
}

\hypersetup{draft}

\begin{document}
\maketitle
\begin{abstract}
We study strategic similarity of game positions in two-player extensive games of perfect information, by looking at the structure of their local game trees, with the aim of improving the performance of game playing agents in detecting forcing continuations. We present a range of measures over the induced game trees and compare them against benchmark problems in chess, observing a promising level of accuracy in matching up trap states.

\end{abstract}

\section{Introduction}

In the field of computer gameplay, Monte Carlo Tree Search (MCTS) agents dominate many games of perfect information such as Go, Checkers, Reversi and Connect Four \cite{Browne:2012}, witness the impressive achievements of the DeepMind team against human players \cite{Silver:2016aa} \cite{alphazero}. Nonetheless, MCTS-based agents still trail behind their $\alpha\beta$ counterparts when playing game positions requiring accurate play. 
 Recent evidence of this fact was provided in game 6 of the 2018 Chess Championship Match between World Champion Magnus Carlsen (White) and challenger Fabiano Caruana (Black), where DeepMind's AlphaZero missed a mate for Black following a sequence of 30 moves, found by the Sesse supercomputer running this line on the non-MTCS-based Stockfish  \cite{gamechanger}.
 
A key issue for MCTS's performance in games like chess is the presence of {\em trap states}, where an initial move may look strong, to then be followed by a forcing sequence by the opponent leading to a loss or significant disadvantage \cite{Ramanujan:2010}. 
Despite the breakthroughs of MCTS-based engines, the challenge still remains to equip MCTS with the capability of handling such forcing lines.

The approach we take in this paper is to formulate a generalised notion of similarity between game states to improve the performance of game playing agents by smart search: if a trap was found after position $x$ and we now analyse position $y$, which is very similar to $x$, then chances are we still have a trap after $y$. As similar strategies are likely to contain similar move sequences but not necessarily similar board positions, our measures are based on possible moves from each position, rather than the appearance of the board itself.

\paragraph{Contribution}
We study similarity of game positions in two-player, deterministic games of perfect information, by looking at the structure of their local game trees, working with the set of possible moves from each position. We  introduce novel similarity measures based on the intersection of move sets, and a structural similarity measure that only considers the arrangement of the local game tree and not the specific moves entailed. We analyse the formal relation of these measures and test them against benchmark problems in chess, with a number of surprising and promising findings. Notably, our structural similarity measure was able to match trap states to their child trap states with  85\% accuracy, without using domain-specific knowledge.
On top of this, we introduce a move matching algorithm, which accurately pairs moves with similar strategic value from different positions. Our results are of immediate relevance to MCTS adaptations to detect and avoid trap states in game play. 

\paragraph{Related Work}

Graph comparison is one of the most fundamental problems of theoretical computer science, with graph isomorphism computation having been an open problem for quite some time \cite{babai}. With tree structures, possibly the most commonly used metric is the {\em edit distance} \cite{Zhang:1989}: based on the number of edits (node insertions, deletions and substitutions) necessary to transform one tree into another, this metric works well for trees of a similar size with many shared nodes and edges. However, it tends to be less suitable when comparing multiple trees of different sizes, as large trees sharing some proportion of their nodes appear further from each other than two completely distinct smaller trees. An alternative measure is the {\em alignment distance} \cite{Zhang:1995}, an adaptation of the edit distance based on the notion of sliding one tree into another and counting the number of edits needed to transform both trees into the combined one. The alignment distance requires lower complexity to compute than the edit distance, but it is technically not a metric and suffers from similar problems comparing trees of different sizes. 

In game playing, the presence of forcing continuations is identified as a key problem faced by AI engines, with more acute implications for chess-like games \cite{Ramanujan:2010}. Surprisingly though, the theory of similarity metrics to aid strategic decisions in game playing is not well developed.


Similarity measures have instead been used in other areas of AI, as in the case of siamese neural networks for one-shot learning \cite{koch2015siamese}. In this case, two symmetric convolutional neural networks were trained on same-different pairs and then shown a test instance as well as one example from each possible classification. The output of the twin networks was then compared using a similarity measure. Here a cross-entropy objective function was used to determine similarity, but this required the networks to be symmetric and weight-tied. New similarity measures based on the structural similarity of networks could remove these requirements, but have not yet been investigated.

\paragraph{Paper Structure}
Section \ref{sec:1} introduces our formal setup to compare game trees through a number of similarity measures. Section \ref{sec:detecting} uses these as the basis of a dynamic algorithm to detect structural similarities among subtrees. In Section 4 we compare these against known chess positions. We conclude discussing potential applications and research directions.

\section{Positional Similarities}\label{sec:1}


Let $G$ be a two-player finite extensive form game of perfect information, where players, e.g., Black and White, alternate moves, with White starting the game. Formally, $G$ consists of a set of histories $(x_0, x_1, \ldots, x_K)$ such that $x_0$ is the starting board position and each $x_{k+1}$ (with $k \leq K$) can be reached from $x_k$ with a single legal move by White, whenever $k$ is even, by Black, otherwise (as in e.g., \cite{maschler}).

We are interested in comparing trees that result from players exploring  game continuations from a certain board position on. In MCTS, for example, these are the game trees generated by the expansion step (see e.g., \cite{suttonbarto}). Let $T_1, T_2$ denote tree roots and $T_{1i}, T_{2i}$ child nodes (board positions) of $T_1$ and $T_2$. Then let $M_i$ denote the set of all possible moves from position $T_i$ and $M_i^d$ the set of all possible moves contained in all possible move sequences of length $d$ from position $T_i$.

We now present three natural measures, of increasing complexity, to establish how similar such trees are: the similarity of continuations, the similarity of sequences and the tree edit similarity. All these measures are model-free, in the sense that they can be used in all situations that can be described as two-player finite extensive games of perfect information. We analyse their formal interrelation in this section and use them as basis of our dynamic algorithm in the subsequent one.

\subsection{Similarity of Continuations} \label{Continuations}
Our first measure, which we call \textit{similarity of continuations}, is calculated from the sets $M_1^d$, $M_2^d$ of 1-ply atomic moves from starting positions $T_1, T_2$ to their children of depth $d$. The similarity is the size of the intersection of these two sets divided by the size of their union.
$$P_{cont}(T_1, T_2) = \frac{| M_1^d \cap M_2^d |}{| M_1^d \cup M_2^d |}$$
As $M_1^d, M_2^d$ can be found from a simple expansion of the game trees, computing such a measure takes time $\mathcal{O}(|M_1^d| + |M_2^d|) \sim \mathcal{O}(b^d)$, where $b$ is the breadth of the game tree. 

At depth 1, the similarity of continuations simply calculates the proportion of children that two nodes share. When extended to a deeper search, the measure becomes less fine-grained since a move that occurs at different depths in the trees will still count as shared, and multiple occurrences of the same move are only counted once.

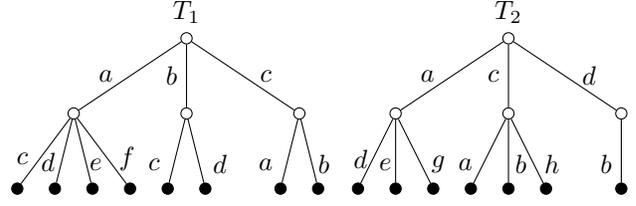
\begin{figure}
\begin{tikzpicture}
    \tikzstyle{level 1}=[level distance=10mm,sibling distance=15mm]
    \tikzstyle{level 2}=[level distance=10mm,sibling distance=5mm]
\tikzstyle{hollow node}=[circle,draw,inner sep=1.5]
\tikzstyle{solid node}=[circle,draw,inner sep=1.5,fill=black]
\node(0)[hollow node,label=above:{$T_1$}]{}
	child{node[hollow node]{} 
		child{node[solid node]{} edge from parent node[left, yshift=-3, xshift=-3]{$c$}}
		child{node[solid node]{} edge from parent node[left, yshift=-4]{$d$}}
		child{node[solid node]{} edge from parent node[right, yshift=-5, xshift=-1]{$e$}}
		child{node[solid node]{} edge from parent node[right, yshift=-3, xshift=3]{$f$}} edge from parent node[left, xshift=-3]{$a$}} 
	child{node[hollow node]{} 
		child{node[solid node]{} edge from parent node[left, yshift=-5, xshift=-3]{$c$}}
		child{node[solid node]{} edge from parent node[right, yshift=-5, xshift=3]{$d$}} edge from parent node[left]{$b$}}
	child{node[hollow node]{} 
		child{node[solid node]{} edge from parent node[left, yshift=-5, xshift=-3]{$a$}}
		child{node[solid node]{} edge from parent node[right, yshift=-5]{$b$}} edge from parent node[right, xshift=3]{$c$}};
\end{tikzpicture}
\begin{tikzpicture}
    \tikzstyle{level 1}=[level distance=10mm,sibling distance=15mm]
    \tikzstyle{level 2}=[level distance=10mm,sibling distance=5mm]
\tikzstyle{hollow node}=[circle,draw,inner sep=1.5]
\tikzstyle{solid node}=[circle,draw,inner sep=1.5,fill=black]
\node(0)[hollow node,label=above:{$T_2$}]{}
	child{node[hollow node]{} 
		child{node[solid node]{} edge from parent node[left, yshift=-3]{$d$}}
		child{node[solid node]{} edge from parent node[left, yshift=-5, xshift=2]{$e$}}
		child{node[solid node]{} edge from parent node[right, yshift=-5, xshift=3]{$g$}} edge from parent node[left, xshift=-3]{$a$}}
	child{node[hollow node]{} 
		child{node[solid node]{} edge from parent node[left, yshift=-5, xshift=-3]{$a$}}
		child{node[solid node]{} edge from parent node[right, yshift=-5, xshift=-1]{$b$}}
		child{node[solid node]{} edge from parent node[right, yshift=-5, xshift=3]{$h$}} edge from parent node[left]{$c$}}
	child{node[hollow node]{} 
		child{node[solid node]{} edge from parent node[left, yshift=-5]{$b$}} edge from parent node[right, xshift=3]{$d$}};
\end{tikzpicture}
\caption{Tree comparison. Starting from the same position, metrics can be used to establish how close the resulting positions will be.}
\label{fig:trees}
\end{figure}

As an example, consider the trees $T_1, T_2$ in Figure \ref{fig:trees}, which have depth-2 continuation sets $M_1^2 = \{a, b, c, d, e, f\}$, \\
$M_2^2 = \{a, b, c, d, e, g, h\}$. 

\smallbreak

Here, 
$$P_{cont}(T_1, T_2) = \frac{| M_1^2 \cap M_2^2 |}{| M_1^2 \cup M_2^2 |} = \frac{5}{8} = 0.625.$$

\subsection{Similarity of Sequences}
Our second similarity measure, which we call \textit{similarity of sequences}, uses longer sequences of moves rather than single plies. To ease computation, we require each possible move sequence of length $d$ from tree root $T_1$ to be first rewritten according to a predetermined move ordering as a \textit{simplified sequence} $S$. Formally, two sequences are simplified into one if and only if they are the same modulo move permutation. These simplified sequences are then stored in a structure $T_1'$, which we call the {\em simplified tree} of $T_1$. As different move permutations can create the same simplified sequence, we also store in $T_1'$ the {\em multiplicity} $k$ of each $S$, where $k$ corresponds to the number of ways $S$ can be reached from the root note. Then the similarity of sequences calculates the ratio of the intersection to the union of the simplified trees. 

Let $k_{1i}$ be the multiplicity of simplified sequence $S_i$ in $T_1'$, $k_{2j}$ the multiplicity of $S_j$ in $T_2'$, and $n = max(|T_1'|, |T_2'|)$, the number of nodes in the larger of $T_1'$, $T_2'$. Then the similarity of sequences of $T_1$ and $T_2$ is given by
$$P_{seq}(T_1, T_2) = \frac{\sum_{i=1}^{n}\sum_{j=1}^{n}(k_{1i} + k_{2j}) \mathds{1}_{S_i \cap S_j}}{\sum_{i=1}^{n} k_{1i} + \sum_{j=1}^{n} k_{2j}}.$$
Calculating the similarity of sequences at depth 2 on the example trees in Figure \ref{fig:trees} is as follows. For an alphabetical ordering, the simplified trees can be written 
$$T_1' = \{ac^2, ad^1, ae^1, af^1, bc^2, bd^1\}$$ 
$$T_2' = \{ac^1, ad^1, ae^1, ag^1, bc^1, bd^1, ch^1\}$$
where the superscript corresponds to the multiplicity of each sequence. Then
$$P_{seq}(T_1, T_2) = \frac{12}{15} = \frac{4}{5} = 0.800.$$
The tree simplification can be done in one depth-first pass of each tree, taking time $\mathcal{O}(b^d)$. Calculating the proportion of shared sequences takes $\mathcal{O}(|T_1'| + |T_2'|)$, which is equal to $\mathcal{O}(b^d)$ in the worst case, the same logarithmic complexity as the similarity of continuations. It should be noted that the tree reduction step means that the complexity coefficient is larger for the similarity of sequences calculation. This is a trade-off for accuracy at depth $d > 1$, as less information is lost when calculating from sequences rather than continuations.

\paragraph{Relation to Kernels}
The similarity of sequences is closely related to the Tanimoto similarity measure or kernel \cite{Swamidass:2005} \cite{Bajusz:2015}, based on the intersection over the union of the inner products of two sets. The Tanimoto kernel was successfully used to calculate the similarity of molecule fingerprints in Bioinformatics from the feature map of a molecule, by counting the number of paths through the map shared by different molecules \cite{Swamidass:2005}. The methods used in this area can be carried over to extensive form games of perfect information, as a board position can be viewed as a fingerprint representing the game that has gone before it. The game tree and feature map can both be traversed and have their matching paths counted. Using a suffix tree data structure \cite{STDS-1} \cite{STDS-2}, we can compute the Tanimoto kernel in time $\mathcal{O}(d(n_1m_1 + n_2m_2))$, for depth $d$, $n_i$ nodes and $m_i$ edges in trees $T_1$, $T_2$. 
The similarity of sequences is also comparable to the random walks kernel \cite{Vishwanathan:2010}, a measure of similarity between two graphs found by counting the number of random paths they share. The main difference here is that the similarity of sequences has limited depth and is a normalised metric.

\subsection{Tree Edit Similarity}
It may be the case that $T_1$ is very similar to $T_2$ but differs by some very shallow moves. If this is the case, the similarity of sequences measure would not detect this similarity. We therefore propose a modified version of the tree edit distance \cite{Zhang:2011}, the \textit{tree edit similarity}, to compare subtrees, which is normalised, and acts as a metric on the tree edit space. The normalised tree edit distance \cite{Zhang:2011} gives values in the range [0,1], and as such would be suitable as a similarity measure when subtracted from 1. The normalised distance is given as
$$\frac{2e(T_1, T_2)}{\alpha(|T_1| + |T_2|) + e(T_1, T_2)},$$
where $e(T_1, T_2)$ is the tree edit distance between $T_1$ and $T_2$, and $\alpha$ is the weight of edit operations. Since there is no need to weight edit operations differently, we may take $\alpha$ to be 1 for all operations. Then, as shown by Li and Zhang, the formula is valid as a metric. Since calculating the distance between two trees is equivalent to calculating their similarity and subtracting it from 1, we define the tree edit similarity as
$$P_{tree}(T_1, T_2) = 1 - \frac{2e(T_1, T_2)}{|T_1| + |T_2| + e(T_1, T_2)}.$$
Calculating the tree edit similarity on the example trees in Figure \ref{fig:trees} is as follows:
$$P_{tree}(T_1, T_2) = 1 - \frac{2 * 6}{11 + 10 + 6} = \frac{15}{27} = 0.556.$$
This measure is the most fine-grained of the three detailed so far. Since calculating tree edit distance on unordered trees is known to be NP-hard \cite{Touzet:2003}, we must again order the nodes in a preprocessing step, with complexity $\mathcal{O}(b^d)$ as above. Once we have ordered trees, the time complexity reduces to $\mathcal{O}(b^{2d}d^2)$ when $d < b$, and $\mathcal{O}(b^{2d+2})$ when $d \geq b$ \cite{Zhang:1989}. As such, the improvements made by the tree edit similarity over the two previous measures must be weighed against the added complexity.

\subsection{Comparing Terminal States}
It may sometimes be necessary to find the similarity of two terminal states. In terms of the game tree for a zero-sum game, two terminal nodes should have a value of 1 if they give the same reward for the agent (win-win, draw-draw, lose-lose), and 0 if the reward is different. Since two terminal nodes have no children, their fractional similarity measure is undefined, so we must handle this case separately. 

The normalised difference between the rewards of the two terminal nodes can be found by subtracting the reward $R_1$ of one node from the reward $R_2$ of the other, then dividing the result by the size of the range of possible reward values $S_0$,  $S_1$. This gives a value between 0 and 1, where 1 represents rewards at opposite ends of the range, and 0 represents equal rewards. Subtracting from 1 then gives a similarity measure, formalised as
$$P(T_1, T_2) = 1 - \frac{| R_2 - R_1 |}{| S_1 - S_0 |}.$$
This can be used in endgame cases to prevent zero errors when calculating other similarity measures. 

\subsection{Relationship Between Measures}
At depth 1, the similarity of sequences and the similarity of continuations are equivalent, as each child move only appears once per tree. At depth 2, the similarity of sequences has greater variation, as can be seen from the following chess-inspired instance.

\begin{example}[Chess trees]
Let $T_1$, $T_2$ be nodes of a chess game tree where branching factor $b$ is constant, and $T_1$, $T_2$ differ only in the placement of two pieces. Then at depth two
$$P_{cont}(T_1, T_2) \sim \frac{b - 2}{b}, \quad P_{seq}(T_1, T_2) \sim \frac{b^2 - 2}{b^2}.$$
Now consider positions $T_3$, $T_4$ that also differ only in the placement of two pieces, except that in $T_3$ the opponent has chosen a forcing move leading to checkmate at depth 2, while in $T_4$ the opponent has chosen otherwise. Then $T_4$ extends past depth two, but $T_3$ is truncated and only contains depth 1 moves, all of which are shared with $T_4$. Then at depth two
$$P_{cont}(T_3, T_4) \sim \frac{b - 2}{b}, \quad P_{seq}(T_3, T_4) \sim \frac{b - 1}{b^2}.$$
So we can see that
$$P_{cont}(T_1, T_2) < P_{seq}(T_1, T_2),$$
$$P_{cont}(T_3, T_4) > P_{seq}(T_3, T_4) $$
and thus the similarity of sequences has greater variation than the similarity of continuations. The tree edit similarity is yet more variable than the similarity of sequences, as can be seen from further calculations on the same examples.
$$P_{tree}(T_1, T_2) \sim \frac{2b^2 - 1}{2b^2 + 1} > P_{seq}(T_1, T_2)$$
$$P_{tree}(T_3, T_4) \sim \frac{1}{2b+1} < P_{seq}(T_3, T_4) $$
\end{example}

Modulo the tradeoff between simplicity and complexity, the above similarity measures can be used to analyse any game trees with a consistent move labelling. This would be especially useful for games with less dynamic trees, that is, those without capturing or blocking moves that change the game tree structurally between plies. For games like Go, with the potential to use one piece to exert power over a whole area, these measures provide useful tools for analysis, which could be further explored by accounting for symmetries and abstractions of the board. 

\section{Detecting Structural Similarities}\label{sec:detecting}
We may find ourselves comparing positions that do not share many continuations, e.g., far away from one another in a game tree. What we can then do is to extend the previous approach to recursively check for subtree similarity.

\begin{algorithm}[t]
\caption{Structural Similarity Detection Protocol}\label{algo3}
\begin{algorithmic}[1]
\Function{StrucSim($T_1$, $T_2$)}{}
\State{$max \gets$ max number of children of $T_1$, $T_2$}
\State{$min \gets$ min number of children of $T_1$, $T_2$}
\For{child $T_{1i}$ of $T_1$}
\State{$sim_1[i] \gets$ \Call{Measure}{$T_{1i}$, $T_1$}}
\EndFor
\For{child $T_{2j}$ of $T_2$}
\State{$sim_2[j] \gets$ \Call{Measure}{$T_{2j}$, $T_2$}} 
\EndFor
\State{pad smaller of $sim_1, sim_2$ with 1s}
\For{$i, j$ from 1 to $max$}
\State{$distances[i, j] \gets | sim_1[i] - sim_2[j] |$}
\EndFor
\State{$matches \gets$ \Call{Match}{$distances$}}
\For{$k$ from 0 to $max$}
\State{$total \gets total + matches[k]$}
\EndFor
\\
\Return{($\frac{distances}{max}$)}
\EndFunction
\end{algorithmic}
\end{algorithm}

\subsection{Structural Similarity Measure}
Our final similarity measure, which we call the \textit{structural similarity measure}, compares the graphical structure of two game trees without comparing their atomic moves directly. The measure is based on calculating the similarity of each starting position $T_1$ to each of its child nodes $T_{1i}$ using any of the three previously defined measures, before comparing this list of similarities to the list of similarities of another starting position $T_2$ to its children $T_{2i}$. The measure uses an assignment algorithm (see Figure \ref{algo3}) to pair each child node of $T_1$ to a child node of $T_2$ to minimise the sum of the paired nodes' similarities to their respective parents. If one subtree has more children than the other, each unpaired child adds 1 to this sum. The sum is then divided by the larger number of children and subtracted from 1 to provide the structural similarity of the two subtrees, where a value of 1 is identical and 0 is completely distinct. 
Let $c_1, c_2$ be the number of child nodes of $T_1, T_2$ respectively. Then, for a selected similarity measure $P$, the structural similarity measure can be expressed as
$$S(T_1, T_2) = 1 - \frac{ | c_1 - c_2 | + \min_{(i, j) \text{pairs}}(\sum | P(T_{1i}) - P(T_{2j}) | )}{\max(c_1, c_2)}$$
The following calculates the structural similarity measure based on the similarity of continuations at depth-1 on the trees in Figure \ref{fig:trees}. The similarity of each branch to its root is
$$P(T_1, T_{1.1}) = \frac{1}{6}, \quad P(T_1, T_{1.2}) = \frac{1}{4}, \quad P(T_1, T_{1.3}) = \frac{2}{3}$$
$$P(T_2, T_{2.1}) = \frac{1}{5}, \quad P(T_2, T_{2.2}) = \frac{1}{5}, \quad P(T_2, T_{2.3}) = 0.$$
There are two minimum distance matchings:
$$\{(T_{1.1}, T_{2.3}), (T_{1.2}, T_{2.1}), (T_{1.3}, T_{2.2})\},$$
$$\{(T_{1.1}, T_{2.3}), (T_{1.2}, T_{2.2}), (T_{1.3}, T_{2.1})\}$$
and their total distance is 0.683. So
$$S(T_1, T_2) = 1 - \frac{0.683}{3} = 0.772.$$
While the structural similarity measure may calculate more accurate similarities between positions, this comes at a cost, as each calculation requires similarity computations of every child node to its parent. When the similarity of sequences or continuations at depth 1 is used as the base measure, on average it takes time $\mathcal{O}(b^2)$ to calculate the similarity of all children to their parent. Assigning children in pairs using the Hungarian algorithm takes $\mathcal{O}(b^3)$ operations, so the structural similarity algorithm runs in time $\mathcal{O}(b^3)$. To improve the complexity, the measure could be approximated by randomly sampling child nodes and calculating their structural similarity, which warrants further investigation.
\paragraph{Move Matching}
As the structural similarity measure pairs moves that are comparably similar to their parent states, this method can be used to pair moves from different board positions that may have similar strategic value. For example, if one position is known to have a killer move in two plies leading to a win for the opponent and this position has a high similarity to a new position, the depth 2 matches can be inspected and the move identified that is most frequently matched to the killer move in the known position, and this move is likely to be a killer move from the new position. We will evaluate the effectiveness of the approach in the forthcoming section.

\paragraph{Generalisability}
The structural similarity measure is generalisable to the analysis of any two local trees with self-consistent move labellings, as the measure can be calculated independently of such labels. This means, e.g., that the structure of a local Go tree can be compared to that of a local chess tree, or, alternatively, we can show how a game tree changes through the game. 

Calculating how dynamic a game is, in terms of the variability of the connection density of the graph, can be very useful in indicating which gameplay heuristics to use. For example, to use the All-Moves-As-First (AMAF) heuristic, which initially updates sibling nodes with the same estimated value for each move played, an agent first assumes that a move from one node is likely to affect the game in a similar way to the same move played from a sibling node. This may be likely to work on less dynamic games but could be less reliable for highly dynamic games, where the effect of a move on the state of the game is less consistent. Conversely, pruning may be most helpful for highly dynamic games, as these offer a stark contrast between reward values for different branches, which is not necessarily the case for less dynamic games.

These hypotheses are supported by studies of successful AMAF use in the less dynamic games Go \cite{Gelly:2011}, Phantom Go \cite{Cazenave:2006}, Havannah \cite{Teytaud:2010} and Morpion Solitaire \cite{Akiyama:2010}, successful pruning in the dynamic game of Amazons \cite{Lorentz:2008} and less successful pruning in Havannah \cite{Teytaud:2010}. 

\section{Performance}\label{sec:results}
We tested how effective the first three similarity measures were at detecting nearby trap states in chess, using the similarity of continuations at depth $d=1$, similarity of sequences at $d=2$ and tree edit similarity at $d=2$. We chose a sample of 4 distinct trap states which each lead to checkmate within 2 to 4 plies, as shown in Figure \ref{fig:chessboards}. We used a sample of all 1000-1500 board positions that were 2 plies away from each trap state, and recorded whether the trap was maintained or not for each new position. The measures were calculated on each of these board positions, as was a cross-correlation measure that was used as a control, calculated by finding the number of squares where piece placement differed and dividing this number by 64. The similarity of sequences was adapted for chess by including captures in the simplified sequences. This adaptation can be generalised to any game with irreversible moves, by recording the irreversible moves from each sequence as well as its standard moves.

\begin{figure}[h]
\chessboard[tinyboard, setfen=rnbqk2r/pppp1ppp/8/2bPp1B1/2P1n3/8/PP2PPPP/RN1QKBNR w KQkq - 3 5, showmover] 
\chessboard[tinyboard, setfen=r2qkb1r/pp2pppp/nnp1P3/4N2b/2P5/7P/PP1N1PP1/R1BQKB1R b KQkq - 2 10, showmover] 
\chessboard[tinyboard, setfen=r1b1k2r/ppppqppp/2n5/4n3/1bP2B2/P4N2/1P1NPPPP/R2QKB1R w KQkq - 0 8, showmover] \chessboard[tinyboard, setfen=r2qkbnr/ppp2ppp/2np4/4N3/2B1P1b1/2N5/PPPP1PPP/R1BQK2R b KQkq - 0 5, showmover] \centering
\caption{Benchmark positions:
(Top Left) Trap from Budapest Gambit: 1.d4  \textsymfigsymbol{N}f6 2. c4 e5 3.d5  \textsymfigsymbol{B}c5 4.\textsymfigsymbol{B}g5  \textsymfigsymbol{N}e4, to be followed by 5.\textsymfigsymbol{B}xd8  \textsymfigsymbol{B}xf2\#.
(Top Right) Trap after mistake in Caro-Kann Defence, Breyer variation: 1. e4 c6 2. d3 d5 3. \textsymfigsymbol{N}d2 dxe4 4. dxe4 \textsymfigsymbol{N}f6 5. \textsymfigsymbol{N}gf3 \textsymfigsymbol{B}g4 6. e5 \textsymfigsymbol{N}d5 7. h3 \textsymfigsymbol{B}h5 8. c4 \textsymfigsymbol{N}b6 9. e6 \textsymfigsymbol{N}a6 10. \textsymfigsymbol{N}e5, to be followed by ... \textsymfigsymbol{B}xd1 11. exf7\#. 
(Bottom Left) Kieninger Trap: 1.d4 \textsymfigsymbol{N}f6 2.c4 e5 3.dxe5 \textsymfigsymbol{N}g4 4.\textsymfigsymbol{B}f4 \textsymfigsymbol{N}c6 5.\textsymfigsymbol{N}f3 \textsymfigsymbol{B}b4+ 6. \textsymfigsymbol{N}bd2 \textsymfigsymbol{Q}e7 7.a3 \textsymfigsymbol{N}gxe5, to be followed by 8.axb4 \textsymfigsymbol{N}d3\#.    (Bottom Right) L\'egal Trap: 1.e4 e5 2. \textsymfigsymbol{N}f3 \textsymfigsymbol{N}c6 3. \textsymfigsymbol{B}c4 d6 4. \textsymfigsymbol{N}c3 \textsymfigsymbol{B}g4 5. \textsymfigsymbol{N}xe5, to be followed by ... \textsymfigsymbol{B}xd1 6. \textsymfigsymbol{B}xf7+ \textsymfigsymbol{K}e7 7. \textsymfigsymbol{N}d5\#.}
\label{fig:chessboards}
\end{figure}

\begin{table}[t]
\begin{center}
\renewcommand{\arraystretch}{1.2}
\resizebox{\columnwidth}{!}{
 \begin{tabular}{llcc} 
\toprule
   \textbf{Position} & \textbf{Similarity Measure} & \textbf{False Negatives} & \textbf{False Positives} \\
\midrule
 {Budapest}  & Continuations & 0.252 & 0.493 \\ 
 \hdashline
  {Caro-Kann} & Continuations & 0.324 & 0.437 \\ 
 \hdashline
  {Kieninger} & Continuations & 0.185 & 0.393 \\ 
 \hdashline
 {L\'egal} & Continuations & 0.332 & 0.461 \\ 
 \hdashline
  {Budapest} & Sequences & 0.234 & 0.480 \\ 
 \hdashline
  {Caro-Kann} & Sequences & 0.335 & 0.397 \\ 
 \hdashline
 {Kieninger} & Sequences & 0.276 & 0.424 \\ 
 \hdashline
 {L\'egal} & Sequences & 0.330 & 0.457 \\ 
 \hdashline
 {Budapest} & Correlation & 0.195 & 0.723 \\ 
 \hdashline
 {Caro-Kann} & Correlation & 0.029 & 0.630 \\ 
 \hdashline
  {Kieninger} & Correlation & 0.865 & 0.380 \\ 
 \hdashline
 {L\'egal} & Correlation & 0.207 & 0.713 \\
\bottomrule
\end{tabular}
}
\caption{Trap states with proportions of false negatives (trap states with lower than average similarity) and false positives (non-trap states with higher than average similarity).}
\label{Threshold_Analysis_Table}
\end{center}
\end{table}

Clearly, an effective measure should evaluate trap states as highly similar to the original position with high frequency, so we fixed a threshold value $\rho$ and calculated the proportion of trap and non-trap states with similarity higher than $\rho$ for each measure. 
For each trap state and each of our similarity measures, when $\rho$ was set to the average value of the similarities, around 70\% of all children which were also trap states had above average similarity to the original position, and consistently over 50\% of non-trap children had below average similarity. This was not the case for the cross-correlation, where up to 87\% of trap states had below average similarity, and 72\% of non-trap states had above average similarity. These results can be seen in Table \ref{Threshold_Analysis_Table}. 

In general, there was no significant difference between the proportion of false positives (non-traps with above average similarity) and false negatives (traps with below average similarity) given by the similarity of sequences, similarity of continuations and tree edit similarity. However the added time complexity of the similarity of sequences and tree edit similarity at depth 2 was significant. Thus, perhaps surprisingly, the similarity of continuations is effectively better as a heuristic similarity measure for evaluating  similarities of closely related board positions than the similarity of sequences.

Finally, for complexity considerations, we tested the structural similarity measure on 5 smaller samples of 40 randomly selected child positions from the first two trap positions. Using this measure, an average of 85\% of child trap states had above average structural similarity to the original position. The high complexity of this measure makes it time-intensive to compute, but results clearly show is rather effective at picking out potential trap states from a select sample of positions.

\paragraph{Move Matching}
The move matching algorithm was also tested on various chess positions, to detect moves with similar strategic impact. Frequent matchings were assumed to be a more reliable indicator of moves with a similar effect on gameplay, so only the top 5 most frequently matched pairings were assessed. 

We tested the matching algorithm on three different samples, each with 6 pairs of board positions, all shown in Table \ref{Matches_Table}. Firstly, we used the algorithm on all traps from the trap detection sample. For all but one of the pairings (L\'egal and Budapest traps), all of the 5 most frequent matches for each pair comprised 2 decisive or 2 non-decisive moves. In all but one pairing (Caro-Kann and Kieninger traps), the two most frequently paired moves were both checkmate moves. 
The second sample we used was based on the L\'egal and Budapest Gambit traps. We compared each trap with a sample of three child positions. This sample comprised one position containing the original trap but a difference in the placement of two pawns; one position where the bishop that had threatened the queen had been captured; and one position that was selected as the best continuation by the Stockfish chess engine. In all but one pairing, all of the top 5 matches comprised 2 decisive or 2 non-decisive moves. All of the most frequently paired moves were both decisive. 
The third sample was a selection of positions from the 2016 World Championship match between Magnus Carlsen and Sergey Karjakin, which appeared after 10, 20, 30 and 40 plies. An average of 4 of the top 5 matches for each pairing comprised 2 decisive or 2 non-decisive moves. Three of the most frequently paired moves were both check moves, and one of them comprised two equivalently unimpactful moves of the king. This sample provided less reliable pairings than the previous two samples, possibly because its positions had a more varied strategic impact than those of the other samples.

These results show that the move matching algorithm is fairly well suited to finding similarly decisive moves from different board positions, and thus useful in detecting possible trap states and sacrificial moves from the game tree structure without evaluating board positions.

\begin{table}[t]
\begin{center}
\renewcommand{\arraystretch}{1.2}
\resizebox{\columnwidth}{!}{
 \begin{tabular}{lcc} 
\toprule
   \textbf{Board Position Pairings} & \textbf{Equally Decisive Top 5 Pairs} & \textbf{Top Match} \\
 \hdashline
 L\'egal, Budapest & 4 & \textsymfigsymbol{B}f7\#, \textsymfigsymbol{B}f2\# \\ 
 \hdashline
  L\'egal, Kieninger & 5 & \textsymfigsymbol{B}f7\#, \textsymfigsymbol{N}f3\# \\ 
  \hdashline
  L\'egal, Caro-Kann & 5 & \textsymfigsymbol{B}f7\#, ef7\# \\ 
 \hdashline
  Budapest, Kieninger & 5 & \textsymfigsymbol{B}f2\#, \textsymfigsymbol{N}d3\# \\ 
 \hdashline
  Budapes, Caro-Kann & 5 & \textsymfigsymbol{B}f2\#, ef7\# \\ 
 \hdashline
  Caro-Kann, Kieninger & 5 & ef7\#, \textsymfigsymbol{N}f3+ \\ 
 \hdashline
  Budapest, Budapest+ 5. b3 a6 & 5 & \textsymfigsymbol{B}f2\#, \textsymfigsymbol{B}f2\# \\ 
 \hdashline
  Budapest, Budapest+ 5. f3 \textsymfigsymbol{N}xe5 & 3 & \textsymfigsymbol{B}f2\#, \textsymfigsymbol{N}f3+ \\ 
 \hdashline
  Budapest, Budapest+ 5. \textsymfigsymbol{B}d2 \textsymfigsymbol{Q}h4 & 5 & \textsymfigsymbol{B}f2\#, \textsymfigsymbol{B}d2+ \\ 
 \hdashline
  {L\'egal}, {L\'egal}+ ... h6 6. a3 & 5 & \textsymfigsymbol{B}f7\#, \textsymfigsymbol{B}f7\# \\ 
 \hdashline
 {L\'egal}, {L\'egal}+ ... h6 6. \textsymfigsymbol{N}xg4 & 5 & \textsymfigsymbol{B}f7\#, \textsymfigsymbol{B}f7+ \\ 
 \hdashline
  {L\'egal}, {L\'egal}+ ... \textsymfigsymbol{N}xe5 6. \textsymfigsymbol{B}e2 & 5 & \textsymfigsymbol{B}f7\#, \textsymfigsymbol{B}b5+ \\ 
 \hdashline
  Carlsen-Karjakin Move 10, Move 20 & 5 & \textsymfigsymbol{B}e6, \textsymfigsymbol{Q}e8\\ 
 \hdashline
  Carlsen-Karjakin Move 10, Move 30 & 5 & \textsymfigsymbol{Q}c3+, \textsymfigsymbol{R}g5+ \\   
 \hdashline
  Carlsen-Karjakin Move 10, Move 40 & 5 & \textsymfigsymbol{Q}c3+, \textsymfigsymbol{R}g5+ \\ 
 \hdashline
  Carlsen-Karjakin Move 20, Move 30 & 4 & \textsymfigsymbol{N}d4, \textsymfigsymbol{R}g5+ \\ 
 \hdashline
  Carlsen-Karjakin Move 20, Move 40 & 3 & \textsymfigsymbol{K}f8, \textsymfigsymbol{K}h8 \\ 
 \hdashline
  Carlsen-Karjakin Move 30, Move 40 & 3 & \textsymfigsymbol{R}g5+, \textsymfigsymbol{Q}g6+ \\ 
\bottomrule
\end{tabular}
}
\caption{Pairs of board positions, the number of equally decisive matches in their 5 most frequent move matches, and their top match.}
\label{Matches_Table}
\end{center}
\end{table}


\section{Applications}
\subsection{AMAF/RAVE Adaptation}
Past papers \cite{Helmbold:2009} have shown that MCTS displays a marked improvement when using adaptations such as All-Moves-As-First (AMAF), Rapid Action Value Estimation (RAVE) and Permutation-AMAF. Such adaptations update multiple areas of the game tree at once, where one move is available from many positions (as in AMAF) or where one board position is a permutation of another, on the assumption that the equivalent move from each of these positions will have the same strategic impact on gameplay. We envisage the effective use of a similarity measure when choosing which equivalent positions to update, as this may lead to more effective trap detection than that of MCTS or its AMAF adaptations. We suggest adding a similarity measure to two MCTS adaptations: the killer heuristic, where decisive moves are evaluated first, and killer RAVE, which only applies RAVE to decisive moves \cite{Lorentz:2011}. MCTS may more quickly detect a trap ahead when combined with these similarity-based adaptations.

\subsection{Wider Game Strategy and Graph Applications}
Many modern AI programs use deep learning to recognise tactical patterns from shapes of features in the field of play. It seems natural to use this learning strategy to group atomic moves by their tactical value, to then create an abstracted game tree with a lower branching factor than the original tree. The structural similarity measure can then be used to detect tactical moves representing equivalent strategies, giving the agent options once it has chosen its desired strategy. 

In cases where an agent is trained to predict the moves a human player would make, as was the case for AlphaGo \cite{Silver:2016aa}, the modified AMAF/RAVE adaptation above can be used to prime the neural network and update predictions for multiple positions at once. This may lead to opportunities for faster reinforcement learning or more efficient learning from smaller data sets. 

\section{Conclusion}
We presented four similarity measures for game positions in two-player, deterministic games of perfect information, based on their game trees with no domain-specific knowledge. We tested the measures on chess and suggested their use in heuristics for MCTS-based agents, noting their application to a range of graphical problems. We showed that, using our first two similarity measures, an average of around 70\% of chess positions occurring 2 plies after a trap state that were also traps had above average similarity to the original position. This figure rose to 85\% using the structural similarity measure. We also showed that our move matching algorithm consistently paired moves with similar strategic value from different starting positions. We believe this can aid MCTS agents in finding equally decisive moves within different areas of the game tree, as well as detecting new trap states.


\bibliographystyle{named}
\bibliography{IJCAI2019Similarity}

\end{document}